\documentclass[runningheads]{llncs}

\usepackage{eccv}

\usepackage{eccvabbrv}

\usepackage{graphicx}
\usepackage{booktabs}

\usepackage[accsupp]{axessibility}  

\usepackage{hyperref}

\usepackage{orcidlink}
\usepackage{multirow}
\usepackage{enumerate}
\usepackage{colortbl}
\usepackage{bbding}
\usepackage[vlined,linesnumbered,ruled]{algorithm2e}
\usepackage{algorithmicx}
\usepackage{algpseudocode}
\usepackage{amsmath}

\makeatletter
\def\thanks#1{\protected@xdef\@thanks{\@thanks
        \protect\footnotetext{#1}}}
\makeatother

\begin{document}

\title{MagDiff: Multi-Alignment Diffusion for High-Fidelity Video Generation and Editing}

\titlerunning{Multi-alignment Diffusion for Video Generation and Editing}

\author{Haoyu Zhao\inst{1,2} \and
Tianyi Lu\inst{1,2} \and
Jiaxi Gu\inst{3} \and
Xing Zhang\inst{1,2} \and 
Qingping Zheng\inst{4} \and
Zuxuan Wu\inst{1,2\dag}\thanks{$^\dag$ Corresponding author} \and
Hang Xu\inst{3} \and
Yu-Gang Jiang\inst{1,2}
}

\authorrunning{Haoyu Zhao et al.}

\institute{Shanghai Key Lab of Intell. Info. Processing, School of CS, Fudan University \and
Shanghai Collaborative Innovation Center on Intelligent Visual Computing \and
\makebox[200pt][l]{Huawei Noah's Ark Lab \and Zhejiang University}
}

\maketitle

\begin{abstract}
  The diffusion model is widely leveraged for either video generation or video editing. As each field has its task-specific problems, it is difficult to merely develop a single diffusion for completing both tasks simultaneously. Video diffusion sorely relying on the text prompt can be adapted to unify the two tasks. However, it lacks a high capability of aligning heterogeneous modalities between text and image, leading to various misalignment problems. In this work, we are the first to propose a unified \textbf{M}ulti-\textbf{a}li\textbf{g}nment \textbf{Diff}usion, dubbed as \textit{MagDiff}, for both tasks of high-fidelity video generation and editing. The proposed MagDiff introduces three types of alignments, including subject-driven alignment, adaptive prompts alignment, and high-fidelity alignment. Particularly, the subject-driven alignment is put forward to trade off the image and text prompts, serving as a unified foundation generative model for both tasks. The adaptive prompts alignment is introduced to emphasize different strengths of homogeneous and heterogeneous alignments by assigning different values of weights to the image and the text prompts. The high-fidelity alignment is developed to further enhance the fidelity of both video generation and editing by taking the subject image as an additional model input. Experimental results on four benchmarks suggest that our method outperforms the previous method on each task.
  \keywords{Video Generation and Editing \and Multi-Alignment Diffusion \and Unified Video Diffusion}
\end{abstract}

\section{Introduction}
\label{sec:intro}

Diffusion model (DM)~\cite{ho2020denoising} has been widely applied to many visual tasks, including video generation and video editing.
Of them, video generation~\cite{ho2022imagen} aims to synthesize a video of good visual quality and video editing~\cite{wu2023tune_a_video} requires the non-edited regions should remain consistent as the source video. Since the two tasks have their task-specific problem, thereby different diffusion models are leveraged to handle them separately. Besides, a group of video editing methods~\cite{wu2023tune_a_video} adopts the one-shot fine-tuning strategy to improve the performance during the inference. Therefore, it is challenging to employ a unified tuning-free diffusion model to support both tasks at the same time.

\begin{figure*}[t]
    \centering
    \captionsetup{type=figure}
    \includegraphics[width=1.0\linewidth]{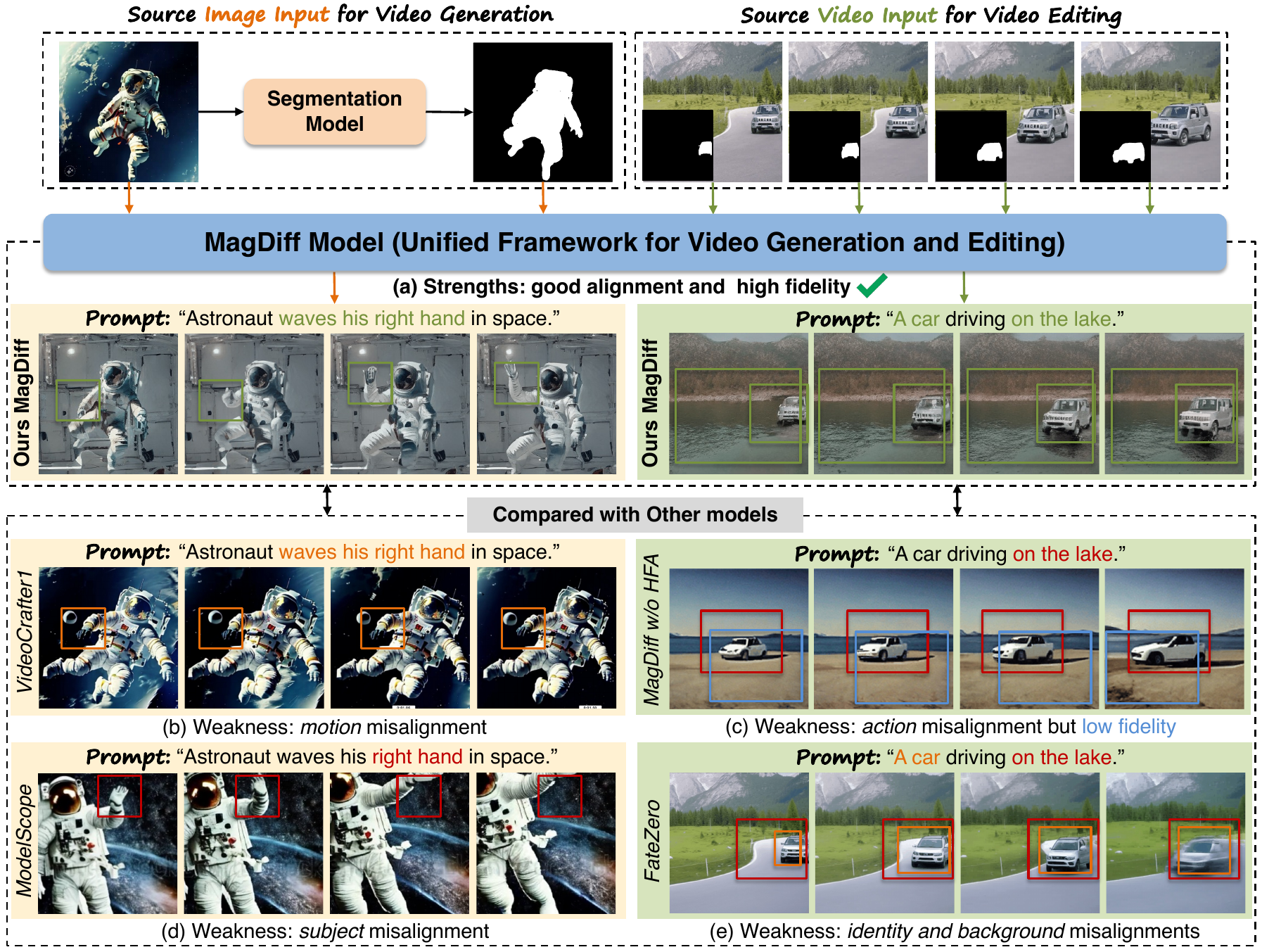}
    \caption{Comparisons of our proposed unified diffusion model named (a) \textit{MagDiff}, (c) MagDiff \textit{w/o} HFA and other video diffusions, including (b) VideoCrafter1~\cite{chen2023videocrafter}, (d) ModelScope~\cite{wang2023modelscope}, and (e) FateZero~\cite{qi2023fatezero}. The results show that our proposed MagDiff obtains the best visual performance (\textit{i.e.} good text-and-image alignment and high fidelity) for both tasks of video generation and editing.}
    \label{fig:task results}
\end{figure*}

Although the traditional video diffusion models~\cite{ho2022imagen, blattmann2023align, ge2023preserve, hong2023cogvideo, he2022lvdm, gu2023reuse} which are only conditioned on the textual prompts can be treated as a framework to adapt both tasks, it lacks a capability of aligning the generated image with the given text prompt.  Specifically, most video generation methods~\cite{singer2022make, ho2022imagen, chen2023videocrafter, ge2023preserve, gu2023reuse} solely relying on textual prompts cannot precisely control the visual details of the synthesized videos because a specific text description normally maps many various videos. For example, given a prompt \textit{``Astronaut waves his right hand in space''}, text-prompt-based video diffusion model suffers from the problem of subject misalignment (see Fig.~\ref{fig:task results}(d) \texttt{the astronaut waves his \textit{left hand} rather than the right hand}). Similarly, existing video editing methods~\cite{qi2023fatezero, chai2023stablevideo, ceylan2023pix2video, liu2023video-p2p, wang2023zero-shot-editing} leverage the text prompt to edit source video raising a more severe problem, such as identity and background misalignments in Fig.~\ref{fig:task results}(e) (editing with tuning-free inference). The main reason is that it is challenging to align heterogeneous modalities of text and image.

To address this issue, an image prompt has been adopted as complementary information by many researchers ~\cite{ni2023conditional, li2023generative, wang2023videocomposer, chen2023videocrafter, zhang2023i2vgen-xl} to control the video generation. These methods can improve the model's ability to align the text prompt and the generated video since the extra image prompt and the generated image are homogeneous. The existing state-off-the-art VideoCrafter1~\cite{chen2023videocrafter} considers both image-and-text prompts as conditions to generate a high-fidelity video with a good subject alignment. However, it is found that employing an image prompt as an additional condition is unable to control the action generation of the subject (also known as ``action misalignment''), like the astronaut in Fig.~\ref{fig:task results}(b) cannot wave his right hand. This is because the VideoCrafter1 method assigns equal weights on both image and text conditions, neglecting the different video controllability between homogeneous and heterogeneous modalities.

To tackle the aforementioned problems, we propose a unified \textbf{M}ulti-\textbf{a}li\textbf{g}nment Diffusion (\textit{MagDiff}) model, a tuning-free method during the inference stage, for high-fidelity video generation and editing. Our proposed MagDiff introduces three various alignments, including Subject-Driven Alignment (SDA) to integrate both tasks in one framework, Adaptive Prompts Alignment (APA) to trade off the controllability between heterogeneous and homogeneous conditions, and High-Fidelity Alignment (HFA) to maintain the fidelity of the subject image. 
Specifically, the SDA segments the subject from an image and employs it as the additional condition rather than the whole image, unifying both tasks of video generation and editing in one model. 
The APA aligns the image prompt and the text prompt with a learnable function in cross-attention blocks, allowing the model to generate more fine-grained content correlated with the subject image and text. 
The HFA aggregates pixel-level multi-scale information into the latent space via a pyramid encoder, reconstructing the visual details of the subject image in the generated videos. 
Experimental results on UCF-101, MSR-VTT, DAVIS, and DreamBooth benchmarks show that our proposed MagDiff achieves good performances on generation and editing tasks in both quantitative and qualitative evaluations.

In summary, our work makes the following contributions:

\begin{itemize}
\item Our \textit{MagDiff} proposes the subject-driven alignment to unify both tasks of video generation and editing in the single framework by using the subject-driven image as an extra condition rather than the full image.

\item Our \textit{MagDiff} develops the adaptive prompts alignment to balance the control strength of homogeneous and heterogeneous conditions, generating fine-grained video well-aligned with both subject image and text.

\item Our \textit{MagDiff} introduces the high-fidelity alignment to improve the high-fidelity of the generated or edited videos by aggregating multi-scale contextual information into the latent space.

\item Experimental results on UCF-101, MSR-VTT, DAVIS, and DreamBooth benchmarks show that our proposed method achieves good results in both quantitative and qualitative evaluations.
\end{itemize}

\section{Related Work}
\label{sec:related-work}

\subsection{Diffusion Models for Video Generation}

The great success of diffusion models in image generation~\cite{dhariwal2021diffusion, ho2022classifier, liu2022compositional, nichol2021improved, preechakul2022diffusion, rombach2022stable_diffusion, betker2023improving} has propelled the advancement of video generation. In the early stage, video generation methods use textual prompts as conditions~\cite{ho2022imagen, blattmann2023align, ge2023preserve, hong2023cogvideo, he2022lvdm, gu2023reuse} to control the synthesized videos. However, conditioning only on textual prompts makes the synthesized videos limited in visual details~\cite{chen2023videocrafter, zhang2023i2vgen-xl, bar2024lumiere}, such as generating a specific subject or background. To this end, recent methods~\cite{ni2023conditional, li2023generative, wang2023videocomposer, girdhar2023emu} draw significant attention to integrating image prompts for video generation. 
The key to image-to-video is adding motion features to the objects in the image. VideoComposer~\cite{wang2023videocomposer} combines the spatial condition (image) and temporal conditions (depth and video) to control the video synthesis. VideoCrafter1~\cite{chen2023videocrafter} takes both the text and image prompts as the inputs and feeds them into the spatial transformer via cross-attention. I2VGen-XL~\cite{zhang2023i2vgen-xl} contains two major stages to get high-resolution videos. The model is trained on large-scale video and image data and then fine-tuned on small high-quality data.
Although current generative models incorporate both images and text as control conditions, existing methods often overlook the differences between these multiple modalities, which will limit controllability in video generation. Therefore, we introduce three novel alignment strategies that take into account these heterogeneous modalities.

\subsection{Diffusion Models for Video Editing}

Before the emergence of text-to-video diffusion models, several studies have explored text-to-image diffusion models for video editing~\cite{qi2023fatezero, wu2023tune_a_video}, with the incorporation of temporal modules to ensure temporal consistency. Tune-A-Video~\cite{wu2023tune_a_video} integrates temporal attention layers into UNet and performs one-shot tuning, while Make-A-Video~\cite{singer2022make} extends the network with spatial-temporal modules to encompass temporal information.
An alternative research direction is influenced by Prompt2Prompt~\cite{hertz2022prompt}  and Plug-and-Play~\cite{tumanyan2023plug}, which enable local editing through attention map manipulation. FateZero~\cite{qi2023fatezero} proposes blending self-attention maps with masks generated by cross-attention maps to facilitate zero-shot video editing. Video-p2p~\cite{liu2023video-p2p} introduces decoupled-guidance attention control to adapt to video scenarios. 
With the development of high-quality text-to-video diffusion models, a line of work employs text-to-video diffusion models for video editing. Dreamix~\cite{molad2023dreamix} introduces a mixed fine-tuning strategy with the Imagen Video model~\cite{ho2022imagen} for better motion editing. Gen-1~\cite{esser2023structure} presents a video diffusion model trained with depth information to govern video structure and content. The popularity of the video diffusion model greatly improves this task.

Nevertheless, the visual clues are still under-explored which is essential for identity-preserving during generation. Literature~\cite{chen2023videocrafter} proposes to preserve the
content of a reference image while generation. However, without carefully considering the weights of image and text prompts, it still falls short in motion editing and identity preserving.

\begin{figure*}[t]
    \centering
    \captionsetup{type=figure}
    \includegraphics[width=1.0\linewidth]{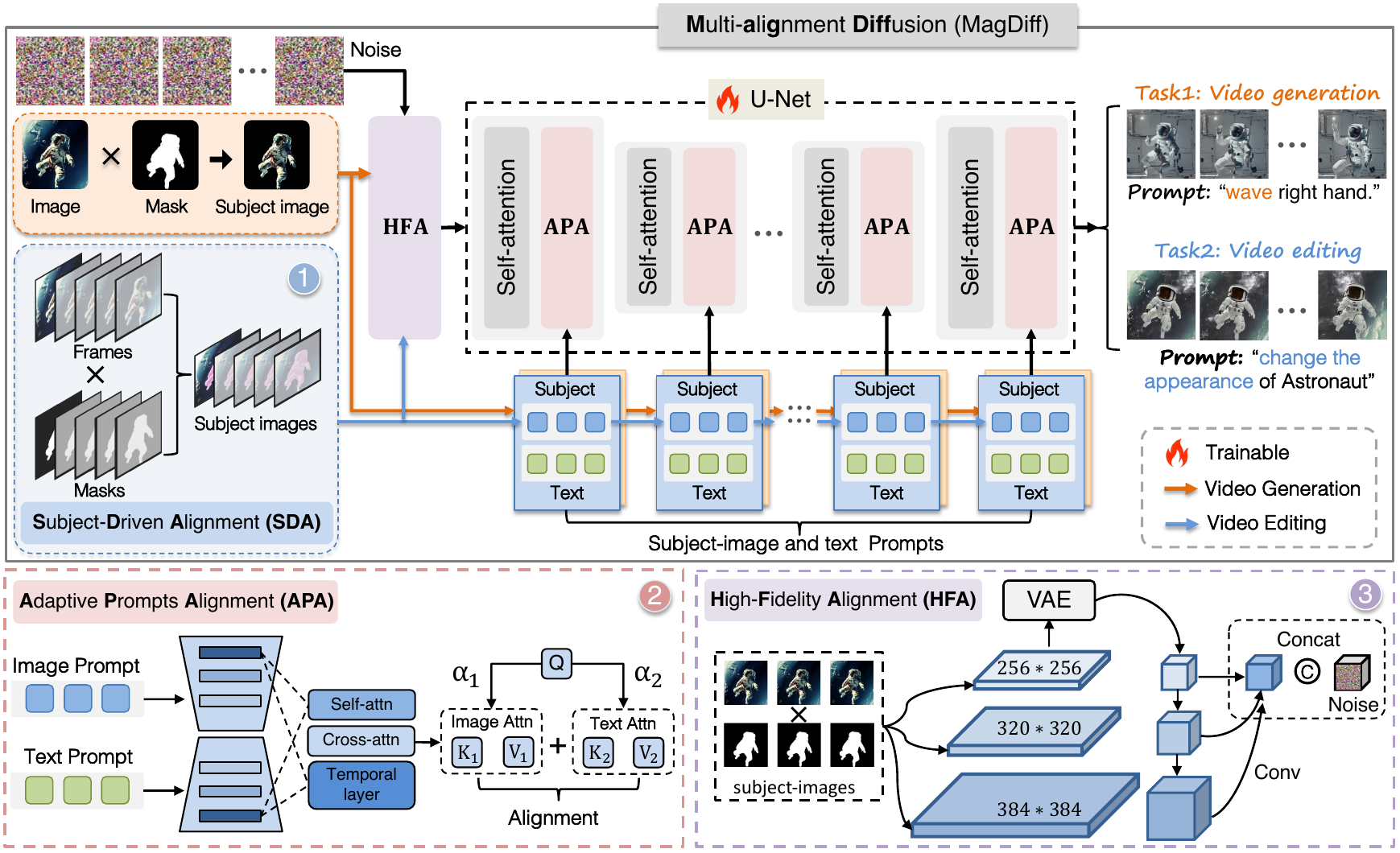}
    \caption{An overview of our proposed \textbf{M}ulti-\textbf{a}li\textbf{g}nment \textbf{Diff}usion (MagDiff), a unified diffusion method supporting both video generation and editing at the same time. Our MagDiff is comprised of three key components: 1) Subject-Driven Alignment (SDA) for unifying two tasks, 2) Adaptive Prompts Alignment (APA) for distinguishing the different controllability between homogeneous and heterogeneous modalities, and 3) High-Fidelity Alignment (HFA) for improving the quality of video generation or editing.
    }
    \label{fig:overview}
\end{figure*}

\section{Multi-alignment Diffusion (MagDiff)}
\label{sec:method}

To achieve high-fidelity video generation and editing tasks in one framework, we propose a \textbf{M}ulti-\textbf{a}li\textbf{g}nment \textbf{Diff}usion (MagDiff) model, which solves the multi-alignments among the generated video, text prompt, and subject-image prompt. Fig.~\ref{fig:overview} shows the overview of our proposed MagDiff conditioned on both text and subject-image prompts to guide video generation and editing using various alignments, including 
Subject-Driven Alignment, Adaptive Prompts Alignment, and High-Fidelity Alignment. The details of each part are depicted below.

\subsection{Subject-Driven Alignment}
\label{sec:subject-alignment}

Video generation and editing are two similar tasks, aiming to generate corresponding content based on text prompts. Differently, the generation task creates a video from pure noise, while the editing task requires the entire video sequence as input and keeps the unchanged parts constant. This difference makes it challenging to unify both tasks in a single model. In this paper, our proposed MagDiff introduces the subject-driven alignment to accommodate both tasks.

\begin{figure*}[t]
    \centering
    \captionsetup{type=figure}
    \includegraphics[width=1.0\linewidth]{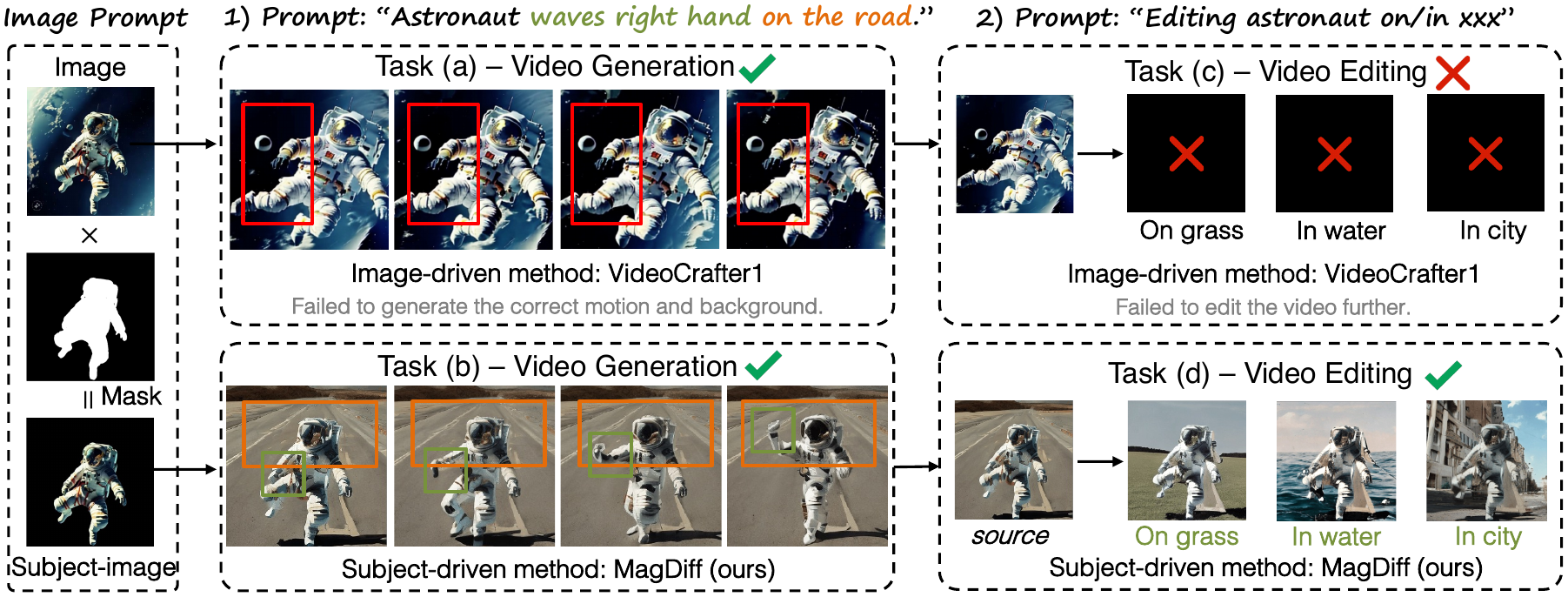}
    \caption{The comparison between the image-driven method (VideoCrafter1~\cite{chen2023videocrafter}) and our subject-driven method MagDiff. The subject-driven method can unify two tasks of video generation and editing but the image-driven method does not have this ability.
    }
    \label{fig:subject-alignment}
\end{figure*}

\paragraph{Image-Driven Alignment.}
Although existing text-to-video generative models~\cite{blattmann2023align, gu2023reuse} can create videos that precisely describe the content of the given prompts, they cannot achieve a customized appearance and keep the identity of the subjects, as shown in Fig.~\ref{fig:task results}(d). Facing these problems, image-driven video diffusions~\cite{chen2023videocrafter,chen2023videodreamer,wang2023videocomposer, zhang2023i2vgen-xl} are proposed to generate videos conditioned on whole reference images and texts. 
Specifically, given the image prompt $\mathbf{c_i}$ and text prompt $\mathbf{c_t}$, the $t$-step denoising process in Diffusion Models (DMs) is denoted as:

\begin{equation}
\label{eq:optimization_func_1}
  \mathbb{E}_{\mathbf{y} \sim \mathcal{N}(\mathbf{0}, \mathbf{I})} [\lVert \mathbf{y} - f_{\theta} (\mathbf{x}_t; \mathbf{c_i},\mathbf{c_t}, t) \rVert_2^2],
\end{equation}

\noindent where the data distribution $p_\text{data}$ is determined by $\mathbf{c_i}$ and $\mathbf{c_t}$ at the same time. However, existing image-driven methods create videos strictly based on the reference images, leading to misalignment between the generated video and text prompts. For instance, given a prompt of ``an astronaut on the road'', Fig.~\ref{fig:subject-alignment}(a) illustrates that  VideoCrafter1~\cite{chen2023videocrafter} can not follow the right prompt. Thereby, the image-driven method cannot edit video content based on the text prompt, failing to unify video generation and editing in one framework.

\paragraph{Subject-Driven Alignment.}
To alleviate this problem, we propose subject-driven alignment to enhance the editability condition on both text and image prompts by balancing the tradeoff between them.
Specifically, we adopt a segmentation method to extract the subject from the full image, obtaining a subject-driven image.
Instead of leveraging the full image as an extra condition, we employ the subject-driven image as the subject-image prompt $\mathbf{c_s}$.
Conditioned on the $\mathbf{c_s}$ and the $\mathbf{c_t}$, our proposed MagDiff can align both the subject image and text prompt with the synthesized videos, unifying tasks of video generation and editing in a single model. Therefore, the denoising process in Eq.~\ref{eq:optimization_func_1} is changed to

\begin{equation}
\label{eq:optimization_func_2}
  \mathbb{E}_{\mathbf{y} \sim \mathcal{N}(\mathbf{0}, \mathbf{I})} [\lVert \mathbf{y} - f_{\theta} (\mathbf{x}_t; \mathbf{c_s},\mathbf{c_t}, t) \rVert_2^2].
\end{equation}

\noindent Through this manipulation, the model can pay more attention to the spatial dimension, not only the temporal information.
As illustrated Fig.~\ref{fig:subject-alignment}(d), utilizing the subject-image prompt can treat the non-editable parts as the ``subject'' and preserve them in the edited videos and utilize text prompt to edit the remaining parts, achieving a unified framework for both tasks. Moreover, to unify the inputs of two tasks, we also propose a standardized data input in latent space, which is introduced in Section~\ref{sec:scale alignment}.

\subsection{Adaptive Prompts Alignment}
\label{sec:prompts alignment}

The subject-image prompt is a homogeneous modality as the generated or edited videos, while the text prompt is a heterogeneous modality. Therefore, the subject image and text prompts have different strengths of alignment to control the video generation in the denoising process.
To make a trade-off between them, we introduce an Adaptive Prompts Alignment (APA) module which can well align both prompts, improving the visual performance.

\begin{figure*}[htbp]
    \centering
    \captionsetup{type=figure}
    \includegraphics[width=1.0\linewidth]{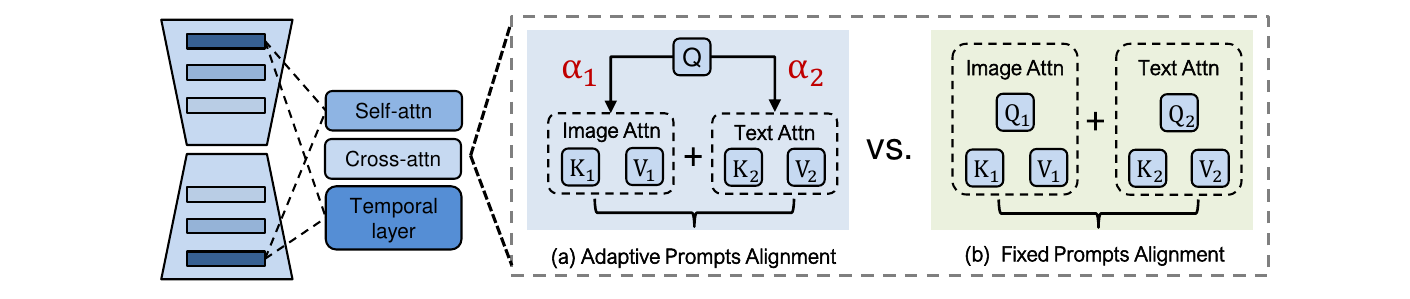}
    \caption{Comparison of our Adaptive Prompts Alignment (APA) vs. Fixed Prompts Alignment. The APA uses two learnable parameters $\alpha_1$ and $\alpha_2$ to adaptively balance the trade-off of alignments between homogeneous and heterogeneous modalities.}
    \label{fig:APA-structure}
\end{figure*}

\paragraph{Fixed Prompts Alignment.} 
To bring the image prompt into the diffusion model, existing frameworks~\cite{ma2023subject_diffusion,molad2023dreamix,chen2023videodreamer} mainly extract the image feature via the CLIP model and directly inject it into the cross-attention structure in U-Net. As described in Fig.~\ref{fig:APA-structure}(b), 
we first use CLIP to encode the text prompt and the subject-image prompt into two feature groups ${Q_1, V_1, K_1}$ and ${Q_2, V_2, K_2}$, separately. Then, we calculate the cross-attention of the two prompts and add them together for feature fusion. The process is denoted as:

\begin{equation}
\label{eq:dual-attention}
\rm{Attention}=\mathrm{Softmax}(\frac{\mathbf{Q_1}\rm{K}_{1}^\top}{\sqrt{d}})\rm{V}_{1}+ \mathrm{Softmax}(\frac{\mathbf{Q_2}\rm{K}_{2}^\top}{\sqrt{d}})\rm{V}_{2}.
\end{equation}

However, such a function allocates equal weights of controllability or editability to both text prompt and image prompt, failing to consider the inherent differences between the two modalities. The misalignment between them leads to generating or editing the low-quality and low-fidelity video since image and text prompts are flexible and variable. 
Besides, prior multi-modal works~\cite{wei2020multi,chen2021crossvit} have found that different cross-attentions bring varying results for specific tasks.

\paragraph{Adaptive Prompts Alignment.} 
To address the homogeneous and heterogeneous modality misalignment, we propose the \textbf{A}daptive \textbf{P}rompts \textbf{A}lignment (APA) module to align the image and text prompts in the denoising process. As shown in Fig.~\ref{fig:APA-structure}(a), in the cross-attention block, we keep $K_1, V_1$ for the text prompt and $K_2, V_2$ for the subject-image prompt, sharing the query $Q$. We then combine the two cross-attention features to obtain the adapted feature.

\begin{equation}
\label{eq:dual-attention}
\rm{Attention}=\alpha_1 * \mathrm{Softmax}(\frac{\mathbf{Q}\mathbf{K}_{1}^\top}{\sqrt{d}})\mathbf{V}_{1}+\alpha_2 * \mathrm{Softmax}(\frac{\mathbf{Q}\mathbf{K}_{2}^\top}{\sqrt{d}})\mathbf{V}_{2},
\end{equation}

\noindent where $\alpha_1, \alpha_2$ are two learnable parameters to dynamically control the visual part and textual part, respectively.
Different from~\cite{chen2023videocrafter}, which only assigns equal weights on both image and text conditions, we find that it is important to give different control for the two modalities.
Through cross-attention reassignment, the adapted feature becomes more semantically correlated with the paired subject-image and text prompts.

Furthermore, we compare the experimental results in Table.~\ref{tab:prompts-alignment-ablation} between the fixed and adaptive prompt alignments and find the APA module has better performance.
To prove such a conclusion, we also visualize the cross-attention map in Fig.~\ref{fig: cross-attention-map}.
It can be found that with the APA module, the refinement is learned after aligning the text prompt and subject-image prompt which can bring more fine-grained generative controllability.

\subsection{High-Fidelity Alignment}
\label{sec:scale alignment}
Although aligning the subject-image prompt and the text prompt can achieve more fine-grained control, the fidelity of the subject image is still overlooked, resulting in a loss of detailed appearances. The inherent reason is that the CLIP model encodes the subject-driven image into the features of the high-dimensional semantics rather than the visual details. To achieve the high fidelity of the subject image, our MagDiff proposes a High-Fidelity Alignment (HFA) module which focuses on aligning the generated video with the subject-image prompt.

Compared with the CLIP encoder~\cite{wang2023videocomposer, ma2023subject_diffusion}, VAE can encode the image to the latent space and decode the latent into the image. Therefore, the HFA shown in Fig.~\ref{fig:overview} is built based on the VAE model and designed as a pyramid structure.  
Specifically, given the noise latent $\mathbf{z}_n$ and the subject image $x_s$, the $x_s$ is sampled into three kinds of sizes. Here, $384 \times 384$, $320 \times 320$ and $256 \times 256$ are chosen empirically. 
Using the VAE model, the subject images with three resolutions are projected into the latent features $\{\mathbf{z}^0_s,\mathbf{z}^1_s,\mathbf{z}^2_s\}$ respectively. 
After passing through the VAE, we use convolutional layers (\textit{convs.}) to align VAE features at three different scales, obtaining the output feature $\mathbf{z}_s$ for the subject image.
Finally, the latent $\mathbf{z_0}$ for denoising is concatenated by $\mathbf{z}_n \oplus \mathbf{z}_s$. 
Such a pyramid structure allows it to accept inputs with different resolutions, which can fit contexts with different scales and increase the robustness of the input.

Additionally, our framework also uses VAE to get the video latent representation. Specifically, we encode the reference image with VAE and don't add noise to it, which can inject the appearance into the denoising step. Notably, the structure of the HFA can unify the inputs for video generation and editing tasks. For generation, we use one subject image while the other images are masked, as shown by the orange line in Fig.~\ref{fig:overview}. For editing, we use all the frames from the original videos as input, which helps maintain the content of the video, as indicated by the blue line in Fig.~\ref{fig:overview}.

\section{Experiments}
\label{sec:experiments}

\subsection{Experimental Setups}

\textbf{Data for Training}. Considering the quality of the video content, we select the Pexel Videos dataset \footnote{https://huggingface.co/datasets/Corran/pexelvideos} to serve as the source data. The dataset contains abundant videos with high quality, each averaging 19.5 seconds in duration. Owing to the lack of subject labels within the original data, we apply the processing approach detailed in \textit{Supplementary} to enhance the dataset's utility for our purposes. We clean up the videos and collect around 76K videos with subjects for training. We initialize the parameters of U-Net from VidRD~\cite{gu2023reuse} (5.3M pre-training data).

\noindent \textbf{Evaluation Datasets}.
We evaluate our MagDiff on four public benchmarks, including UCF-101, MSR-VTT, DreamBooth, and DAVIS. UCF-101~\cite{soomro2012ucf101} has 101 brief class names (10,000 videos for test), which is commonly employed to assess the generation performance of various methods~\cite{singer2022make, hong2023cogvideo, wang2023videofactory}. MSR-VTT~\cite{xu2016msrvtt} contains 2,990 videos for testing. DAVIS~\cite{jay2023loveu} dataset is proposed for video editing task and DreamBooth~\cite{ruiz2023dreambooth} dataset is built to evaluate the fidelity of the subject.

\noindent
\textbf{Evaluation Metrics.}
We mainly use two aspects of evaluation metrics:

(i) Metrics for video quality evaluation. Previous works like~\cite{singer2022make, ge2023preserve, blattmann2023align} use two metrics for quantitative evaluation, i.e., \textbf{Fr\'{e}chet Video Distance (FVD)} ~\cite{thomas2019fvd} and Video \textbf{Inception Score (IS) }~\cite{saito2020generate}. FVD is a video quality evaluation metric based on FID~\cite{parmar2022fid}. Following~\cite{singer2022make}, we use a trained I3D model~\cite{carreira2017i3d} for calculating FVD. Following previous works ~\cite{singer2022make, hong2023cogvideo, blattmann2023align}, a trained C3D model~\cite{tran2015leanring} is used for calculating the video version of IS.

(ii) Metrics for identity consistency and video-prompt alignment.
a) We compute the \textbf{DINO score}~\cite{ruiz2023dreambooth} between the generated subject and the given subject image to evaluate the fidelity. b) Following ~\cite{wu2023tune_a_video}, we calculate the average cosine similarity between all pairs of video frames to evaluate the \textbf{Frame-consistency}. We calculate the average CLIP score between all frames of generated videos and corresponding prompts to evaluate the \textbf{Textual-alignment}.

\subsection{Comparison with State-of-the-Art}
To evaluate the effectiveness of our unified MagDiff on both video generation and editing, we conducted comparative analyses with various state-of-the-art methods for two tasks separately. For video generation, we compare our method with the existing video generation methods including text-to-video and text\&image-to-video. For video editing, we compare our method with the current video editing methods containing tuning-free and fine-tuning ways during inference.

\begin{table}[ht]
    \caption{
    \textbf{For video generation}, quantitative comparison of MagDiff and other methods on UCF-101 and MSR-VTT. All the videos are generated in a zero-shot manner. 
    }
    \centering
    \small
    \resizebox{\linewidth}{!}{
    \begin{tabular}{lclclcrrcr}
       \toprule
       \multirow{2}{*}{\textbf{Models}}  &&\multirow{2}{*}{\textbf{Input Type}} &&\textbf{Training} &&\multicolumn{2}{c}{\textbf{UCF-101}}  && \textbf{MSR-VTT} \\
        \cmidrule{7-8} \cmidrule{10-10}
        && &&\textbf{Videos} && IS $\uparrow$ & FVD $\downarrow$ && FVD $\downarrow$ \\
       \midrule
       LVDM~\cite{he2022lvdm} && text to video && 2.0M && - & 641.80 && - \\
       ModelScope~\cite{wang2023modelscope} && text-to-video && 10M && - & - && 550 \\
       Make-A-Video~\cite{singer2022make} && text-to-video && 20.0M && 33.00 & 367.23 && -\\
       VideoFactory~\cite{wang2023videofactory} && text-to-video && 140.7M && - & 410.00 && -\\
       PYoCo~\cite{ge2023preserve}  && text-to-video && 22.5M && 47.76 & 355.19 && -\\
         \cmidrule{1-1} \cmidrule{3-3} \cmidrule{5-5} \cmidrule{7-8} \cmidrule{10-10}
       I2VGen-XL~\cite{zhang2023i2vgen-xl} && text\&image-to-video && 10M && 18.90 & 597.49 && -\\
       VideoComposer~\cite{wang2023videocomposer} && text\&image-to-video && 10.3M && - & - && 580\\
       VideoCrafter1~\cite{chen2023videocrafter} && text\&image-to-video && 10.3M && 44.53 & 415.87 && 465\\
       \rowcolor{gray!20}
       \textbf{MagDiff (Ours)} && text\&image-to-video && \textbf{5.3M+76K} && \textbf{48.57} & \textbf{339.62} && \textbf{245}\\
       \bottomrule
    \end{tabular}}
    \label{tab:eval_ucf101}
\end{table}

\begin{table}[ht]
    \caption{\textbf{For video generation}, quantitative comparison of our MagDiff and other methods from the additional metrics on UCF-101, MSR-VTT, and DreamBooth. 
    }
    \centering
    \small
    \resizebox{\linewidth}{!}{
    \begin{tabular}{lcccccccccccc}
       \toprule
       \multirow{3}{*}{\textbf{Methods}}       && \multicolumn{3}{c}{\textbf{UCF-101}} && \multicolumn{3}{c}{\textbf{MSR-VTT}} && \multicolumn{3}{c}{\textbf{DreamBooth}}\\
       \cmidrule{3-5} \cmidrule{7-9} \cmidrule{11-13}
       && \multirow{2}{*}{DINO} & Textual & Frame 
       && \multirow{2}{*}{DINO} & Textual & Frame 
       && \multirow{2}{*}{DINO} & Textual & Frame\\
       & && align & consist & && align & consist & && align & consist\\
       \cmidrule{1-1} \cmidrule{3-13}
       AnimateDiff (V3)~\cite{guo2023animatediff} && 46.2 & 24.1 & 89.8 
                                                  && 47.3 & 22.4 & 89.6 
                                                  &&59.1 &23.2 &90.4 \\
          I2VGen-XL~\cite{zhang2023i2vgen-xl}     &&44.1 &21.3 &89.4 
                                                  &&42.7 &18.4 &\textbf{89.8} 
                                                  &&58.6 &22.9 &89.7 \\
      \rowcolor{gray!20}
         \textbf{MagDiff (Ours)} &&\textbf{50.8} &\textbf{25.4} &\textbf{90.2}
                                                   &&\textbf{50.2} &\textbf{23.2} &88.4
                                                   &&\textbf{61.4} &\textbf{25.4} &\textbf{92.2}  \\
       \bottomrule
    \end{tabular}}
    \label{tab:dino_alignment_consistency}
\end{table}

\noindent
\textbf{Evaluation of Video Generation.}
Table~\ref{tab:eval_ucf101} exhibits the best performance of our proposed MagDiff in terms of all metrics on the UCF-101 and MSR-VTT benchmark datasets.
Specifically, our method surpasses all text-to-video generation methods, since it utilizes the subject within the image, which contains less visual information.
Besides, our MagDiff also outperforms other text\&image-to-video methods on both FVD and IS metrics, because we use the subject-driven image as a condition rather than the full image.
To demonstrate the fidelity and identity consistency of subject images within the videos synthesized by MagDiff, we employ the DINO score, Textual-align, and Frame-consistency for evaluative purposes on the UCF-101, MSR-VTT, and DreamBooth compared with~\cite{guo2023animatediff,zhang2023i2vgen-xl}.
The results in Table~\ref{tab:dino_alignment_consistency} indicate that the videos we generate have good text alignments, and the continuity between frames is also commendable.

\noindent
\textbf{Evaluation of Video Editing.} 
Table~\ref{tab:video editing} compares our MagDiff and other video editing methods on the popular DAVIS dataset with CLIP-Score metrics of text and image. The results show that our MagDiff surpasses the tuning-free method Framewise IP2P~\cite{brooks2023instructpix2pix}, with improvements of 2.54 and 4.10 on the textual-align and frame-consistency metrics respectively.  Compared to fine-tuning methods during inference, our tuning-free MagDiff method is much more robust and performs competitively with the Tune-A-Video~\cite{wu2023tune_a_video} and FateZero~\cite{qi2023fatezero} methods, verifying the superiority of our proposed MagDiff.

\begin{table}[ht]
    \setlength{\tabcolsep}{1.5mm}
    \caption{\textbf{For video editing}, quantitative comparison of our MagDiff and other video editing methods on DAVIS benchmark.
    }
    \centering
    \small
    \resizebox{\linewidth}{!}{
    \begin{tabular}{lcccccc}
       \toprule
       \textbf{Methods} &&  \textbf{Inference method} &&  \textbf{Textual-align} &&  \textbf{Frame-consistency} \\
       \cmidrule{1-1} \cmidrule{3-3} \cmidrule{5-5} \cmidrule{7-7}
       Tune-A-Video~\cite{wu2023tune_a_video} &&Fine-Tuning  && 28.33 &&  90.45 \\
       FateZero~\cite{qi2023fatezero}  &&Fine-Tuning  && 23.81  && 92.92 \\
       \midrule
       Framewise IP2P~\cite{brooks2023instructpix2pix}  &&Tuning-Free  && 25.11  && 86.76 \\
       \rowcolor{gray!20}
       \textbf{MagDiff (Ours)}   &&Tuning-Free && \textbf{27.65} (+2.54)  && \textbf{90.86} (+4.10)  \\
       \bottomrule
    \end{tabular}}
    \label{tab:video editing}
\end{table}

\noindent
\textbf{Human Evaluations.} We perform human evaluations by using a panel of 34 human raters, over 15 videos with corresponding prompts for each method. We adopt the Likert Scale~\cite{likert1932technique} to evaluate subject fidelity, prompt alignment, and the quality of the generated videos. The range of these three scores is between 1 and 5, which represents from very dissatisfied to very satisfied. Table~\ref{tab:human_evaluation} compares our MagDiff with both VideoCrafter1~\cite{chen2023videocrafter} and AnimateDiff (V3)~\cite{guo2023animatediff} on human evaluations. We maintain the original input format of these models, which is the complete image without the mask.
Our method achieves the highest scores on both subject fidelity and text alignment, as well as producing high-quality videos. 
These results indicate that our method performs well under human evaluation and demonstrates its superiority over existing methods.

\begin{table}[ht]
    \caption{Human-preference aligned results from three different aspects, with the rank of each aspect in the brackets.}
    \centering
    \small
    \resizebox{\linewidth}{!}{
    \begin{tabular}{lccccccc}
       \toprule
       \textbf{Methods} && \textbf{Image-prompt Alignment} && \textbf{Text-prompt Alignment} && \textbf{Quality} \\
       \cmidrule{1-1} \cmidrule{3-3} \cmidrule{5-5} \cmidrule{7-7}
       VideoCrafter1~\cite{chen2023videocrafter} && 3.2  && 2.8  && 3.2 \\
       AnimateDiff (V3)~\cite{guo2023animatediff} && 3.5  && 3.6  && 3.3 \\
              \rowcolor{gray!20}
       \textbf{MagDiff (Ours)} && \textbf{4.4}  && \textbf{4.1}  && \textbf{3.7} \\
       \bottomrule
    \end{tabular}}
    \label{tab:human_evaluation}
\end{table}

\begin{figure*}[t]
\centering
\includegraphics[width=0.90\linewidth]{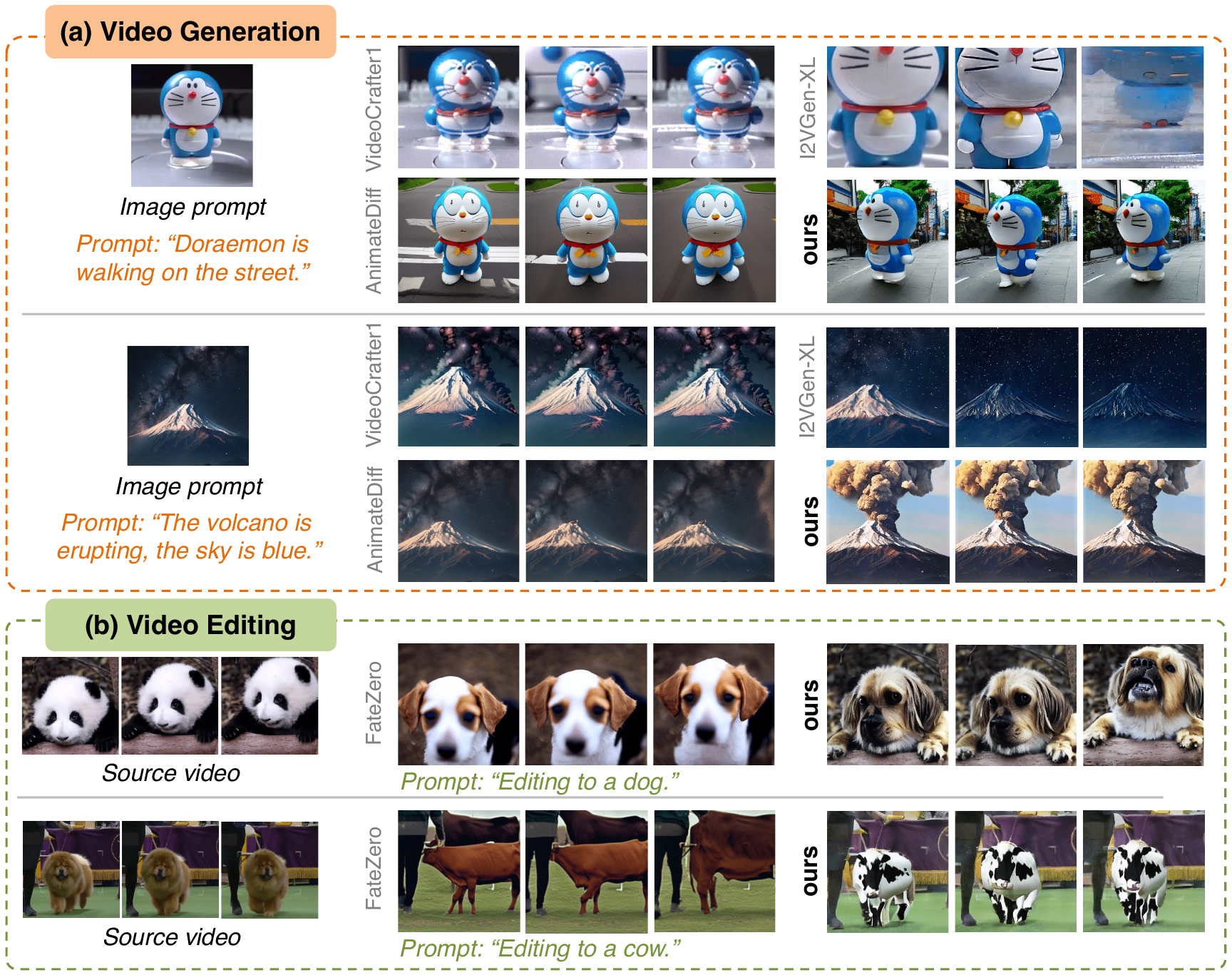}
\caption{For qualitative evaluation, we compare our MagDiff with VideoCrafter1~\cite{chen2023videocrafter}, I2VGen-XL~\cite{zhang2023i2vgen-xl}, and AnimateDiff (V3)~\cite{guo2023animatediff} on the generation task (orange dotted box) and compare with FateZero~\cite{qi2023fatezero} on the editing task (green dotted box).
}
\label{fig: visual-results-compared}
\end{figure*}

\noindent
\textbf{Qualitative Evaluation.}
Fig.~\ref{fig: visual-results-compared} offers a visual comparison between our MagDiff model and other methods for both tuning-free video generation and editing.
Examples in Fig.~\ref{fig: visual-results-compared}(a) present that our method can synthesize smoother and higher fidelity videos than other video generation methods. Fig.~\ref{fig: visual-results-compared}(b) demonstrates that our method can edit video contents more aligned with the user-defined text prompts. This further verifies the effectiveness of our unified framework.

\subsection{Ablation Studies}

\begin{table}[h]
  \setlength{\tabcolsep}{3.5mm}
    \caption{\textbf{Baseline analysis of SDA, APA, and HFA within MagDiff}. We use the VidRD~\cite{gu2023reuse} as the basemodel, which is a text-to-video model.}
    \centering
    \small
    \resizebox{\linewidth}{!}{
    \begin{tabular}{lcrcrcrcrcrcr}
       \toprule
       \multirow{2}{*}{\textbf{Methods}} &&\multirow{2}{*}{\textbf{HFA}} &&\multirow{2}{*}{\textbf{APA}} &&\multirow{2}{*}{\textbf{SDA}} && \multicolumn{3}{c}{\textbf{UCF-101}}&&\textbf{MSR-VTT} \\
             \cmidrule{9-11} \cmidrule{13-13}
        && && &&  && \textbf{IS $\uparrow$} && \textbf{FVD $\downarrow$} && \textbf{FVD $\downarrow$}\\
       \midrule
       \multirow{6}{*}{MagDiff} &&\Checkmark  &&\XSolidBrush &&  \XSolidBrush  && 42.11 && 530.26 && 372 \\
       &&\XSolidBrush &&\Checkmark && \XSolidBrush && 40.47 && 534.84 && 394 \\
        &&\Checkmark &&\Checkmark && \XSolidBrush && 43.39 && 444.67 && 311 \\
        &&\Checkmark &&\XSolidBrush && \Checkmark  && 45.74 && 367.23 && 274 \\
        &&\XSolidBrush &&\Checkmark && \Checkmark  && 46.85 && 388.41 && 286 \\
       \rowcolor{gray!20}
        &&\Checkmark &&\Checkmark && \Checkmark  && \textbf{48.57} && \textbf{339.62} && \textbf{245} \\
       \bottomrule
    \end{tabular}}
    \label{tab:decoup-evaluation}
\end{table}

\noindent
\textbf{Baseline analysis of three alignments.}
Table~\ref{tab:decoup-evaluation} analyzes the influences of the SDA, APA, and HFA modules for our MagDiff. The base model is the VidRD~\cite{gu2023reuse}, which is a T2V method.
We find that using the SDA can enhance the overall effect of our method. We analyze that the mask can help the model better implement text control capabilities. Meanwhile, we suppose that the HFA can introduce specific pixel information directly into the latent space, it also plays a significant role in improving the effectiveness of the model.

\begin{figure*}[t]
\centering
\includegraphics[width=0.95\linewidth]{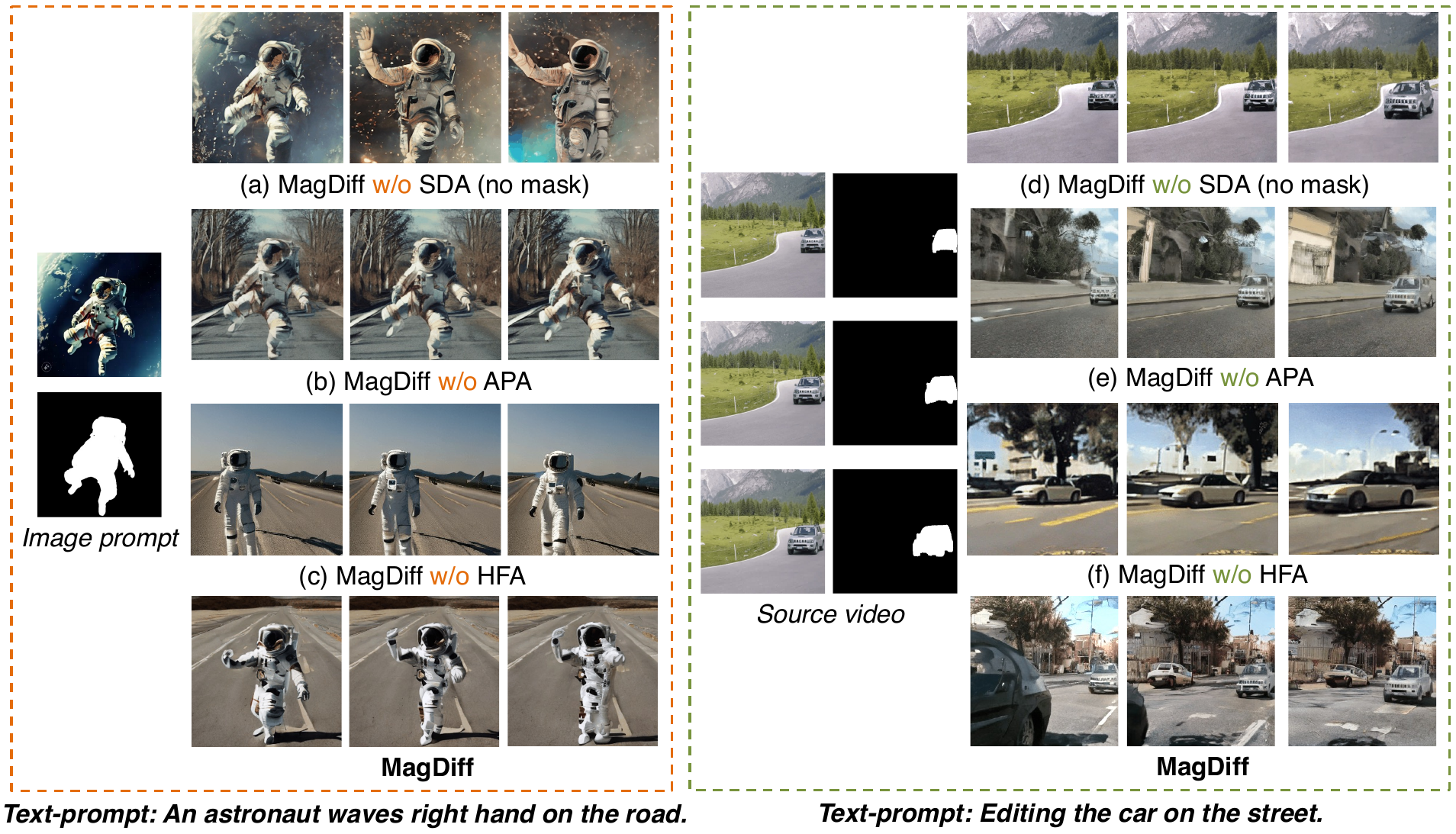}
\caption{Visualization of the three alignments in MagDiff, including SDA, APA, and HFA. The orange and green dotted boxes show the generated and edited results.}
\label{fig: show-case-ablation}
\end{figure*}

Additionally, as presented in Fig.~\ref{fig: show-case-ablation}, \textbf{1)} compared to the first and last rows in Fig.~\ref{fig: show-case-ablation}, it can be observed that lacking subject-driven prompts generates videos unmatching the user-defined text prompt (see Fig.~\ref{fig: show-case-ablation}(a)) and is also unable to edit the source videos conditioned on the given text description (Fig.~\ref{fig: show-case-ablation}(d)). This validates the effectiveness of our proposed SDA. Besides, \textbf{2)} compared to the second and last rows in Fig.~\ref{fig: show-case-ablation}, we find that while missing APA module in MagDiff can generate or edit a video, better aligning the text condition than the first rows, it suffers from a low-fidelity of visual results. This is due to overlook the fidelity of the subject image. Consequently, \textbf{3)} compared to the last two rows, it can be found that adding the HFA is useful to improve the video's fidelity because the pixel-level multiscale information is aggregated into the latent space.

\begin{figure*}[t]
\centering
\includegraphics[width=1.0\linewidth]{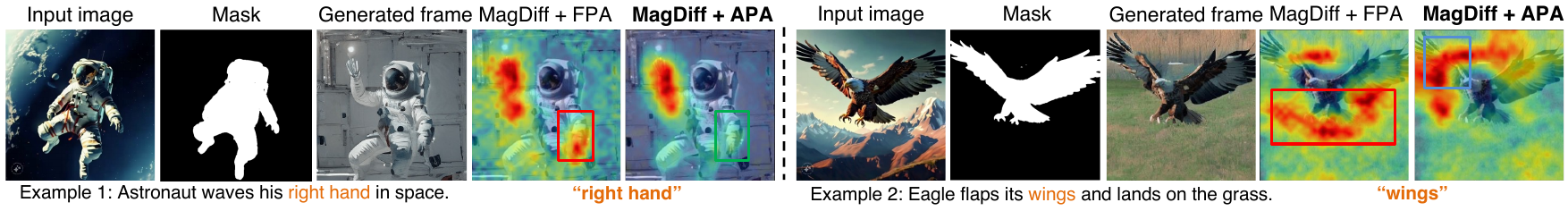}
\caption{Visualization of cross-attention maps between the ``MagDiff +'' FPA and APA.}
\label{fig: cross-attention-map}
\end{figure*}

\noindent
\textbf{Effect of APA.} 
To validate the effectiveness of our proposed APA, we compare it with the Fixed Prompts Alignment (FPA). 
Table.~\ref{tab:prompts-alignment-ablation} illustrates that ``MagDiff w/ APA'' outperforms ``MagDiff w/ FPA'' by a clear margin. 
On the UCF-101 and MSR-VTT datasets, ``MagDiff w/ APA'' has FVD scores of 48.79 and 46 lower than ``MagDiff w/ FPA'', respectively.
The results show the superior performance of our proposed APA, verifying that better alignment between the subject-image prompt and the text prompt can help the model understand fine-grained text. 
To further prove this, in Fig.~\ref{fig: cross-attention-map}, conditioned on the text prompts, we visualize the APA's cross-attention maps between the generated video frame and the words ``right hand'', and ``wings''. Results demonstrate that FPA wrongly activates both the astronaut's left and right hands, while APA has a better understanding and activates the right hand corresponding to the text description. 

\begin{table}[t]
  \setlength{\tabcolsep}{5.0mm}
    \caption{Performance comparison of FPA and APA within MagDiff. ``MagDiff w/ APA'' denotes the use of APA, while ``MagDiff w/ FPA'' denotes the use of FPA.}
    \centering
    \small
    \resizebox{\linewidth}{!}{
    \begin{tabular}{lcrcrcr}
       \toprule
       \multirow{2}{*}{\textbf{Methods}} && \multicolumn{3}{c}{\textbf{UCF-101}}&&\textbf{MSR-VTT} \\
             \cmidrule{3-5} \cmidrule{7-7}
        &&   \textbf{IS $\uparrow$} && \textbf{FVD $\downarrow$} && \textbf{FVD $\downarrow$}\\
       \midrule
       MagDiff w/o FPA && 44.26 && 388.41 && 291 \\
       \rowcolor{gray!20}
       \textbf{MagDiff w/ APA} && \textbf{48.57} && \textbf{339.62} && \textbf{245} \\
       \bottomrule
    \end{tabular}}
    \label{tab:prompts-alignment-ablation}
\end{table}

\noindent
\textbf{Effect of different values for $\alpha_1$ and $\alpha_2$.} 
We also explore the impact of different values for $\alpha_1$ and $\alpha_2$ on the APA module. Table~\ref{tab:ablation_alpha} compares the effects of fixing the two parameters with trainable. We conduct comparisons for three different group values of $\alpha_1$ and $\alpha_2$. The results show that the model achieves optimal performance when using a trainable way. This is because trainable parameters better balance image-text during training, improving model performance.

\begin{table}[ht]
    \setlength{\tabcolsep}{5.0mm}
    \caption{Effects of different values for \(\alpha_1\) and \(\alpha_2\) in APA cross-attention.}
    \centering
        \small
    \resizebox{\linewidth}{!}{
    \begin{tabular}{lcrcrcr}
       \toprule
       \multirow{2}{*}{\textbf{Methods}} &&\multicolumn{3}{c}{\textbf{UCF-101}}&&\textbf{SR-VTT} \\
             \cmidrule{3-5} \cmidrule{7-7}
        &&  \textbf{IS $\uparrow$} && \textbf{FVD $\downarrow$} && \textbf{FVD $\downarrow$}\\

       \midrule
       $\alpha_1=0.3$ \& $\alpha_2=0.7$ && 46.45 && 391.72 && 283 \\
       $\alpha_1=0.5$ \& $\alpha_2=0.5$ && 45.11 && 364.89 && 296 \\
       $\alpha_1=0.7$ \& $\alpha_2=0.3$ && 47.20 && 386.32 && 267 \\
                     \rowcolor{gray!20}
       \textbf{Trainable} && \textbf{48.57} && \textbf{339.62} && \textbf{245} \\
       \bottomrule
    \end{tabular}}
    \label{tab:ablation_alpha}
\end{table}

\section{Conclusion}
In this work, we first propose a unified diffusion model named MagDiff, for both the video generation and editing tasks. In our framework, we mainly solve the three kinds of alignments to achieve high-fidelity video generation, including subject-driven alignment, adaptive prompts alignment, and high-fidelity alignment. Experimental results based on four benchmarks show that our method achieves good results in both quantitative and qualitative evaluations.

\section*{Acknowledgements} 
This project was supported by NSFC under Grant No. 62032006  and No. 62102092. Qingping Zheng was supported by the Postdoctoral Fellowship Program of CPSF under Grant No. GZC20241489.

%
%
\bibliographystyle{splncs04}
\bibliography{main}

\newpage

\title{MagDiff: Multi-Alignment Diffusion for High-Fidelity Video Generation and Editing
\\ --- Supplementary Material ---}

\titlerunning{Multi-alignment Diffusion for Video Generation and Editing}

\author{Haoyu Zhao\inst{1,2} \and
Tianyi Lu\inst{1,2} \and
Jiaxi Gu\inst{3} \and
Xing Zhang\inst{1,2} \and 
Qingping Zheng\inst{4} \and
Zuxuan Wu\inst{1,2\dag}\thanks{$^\dag$ Corresponding author} \and
Hang Xu\inst{3} \and
Yu-Gang Jiang\inst{1,2}
}

\authorrunning{Haoyu Zhao et al.}

\institute{Shanghai Key Lab of Intell. Info. Processing, School of CS, Fudan University \and
Shanghai Collaborative Innovation Center on Intelligent Visual Computing \and
\makebox[200pt][l]{Huawei Noah's Ark Lab \and Zhejiang University}
}

\maketitle

\renewcommand{\thetable}{\Alph{table}}
\renewcommand{\thefigure}{\Alph{figure}}
\renewcommand\thesection{\Alph{section}}

\section{Overview}
\label{sec: overview}
We provide more details about MagDiff including:

\begin{itemize}
\item We demonstrate the training details of MagDiff in Section~\ref{sec: training-details}.

\item We introduce our method of training data process in Section~\ref{sec: data process}, and the details of the model and experiments in Section~\ref{sec: model and experiments details}.

\item We provide more qualitative results of video generation and editing in Section~\ref{sec: supple_more_visualizations}.

\item We conclude the limitation in Section~\ref{sec: supple_limitation}.
\end{itemize}

\section{Training Details}
\label{sec: training-details}
During the model training, we sample eight frames for one video to train the denoising network.
Our MagDiff is initialized with weights from VidRD model~\cite{gu2023reuse}. For model training, we utilize high-quality data from existing video dataset.
For the frame input, we utilize the widely used data augmentation strategy for training, including center crop and random shuffle. We also set a 15\% probability of randomly dropping the prompt during training.
Noticed that we do not train all the parameters, only the temporal layers, transformer blocks of the spatial layers, and the newly added projections are trainable.
The learning rate is set at $\rm{5 \times 10^{-5}}$ for all training tasks.
The VAE model and CLIP model are frozen.
When doing inference, the classifier-free guidance is set as 7.5. For latent diffusion model sampling, we use DDIM in all our experiments.
All experiments are conducted using eight Nvidia Tesla V100 GPUs for 50\textit{K} iterations with a batch size of 64.

\section{Data Process for Training}
\label{sec: data process}

\begin{figure}[h]
    \centering
    \captionsetup{type=figure}
    \includegraphics[width=0.7\linewidth]{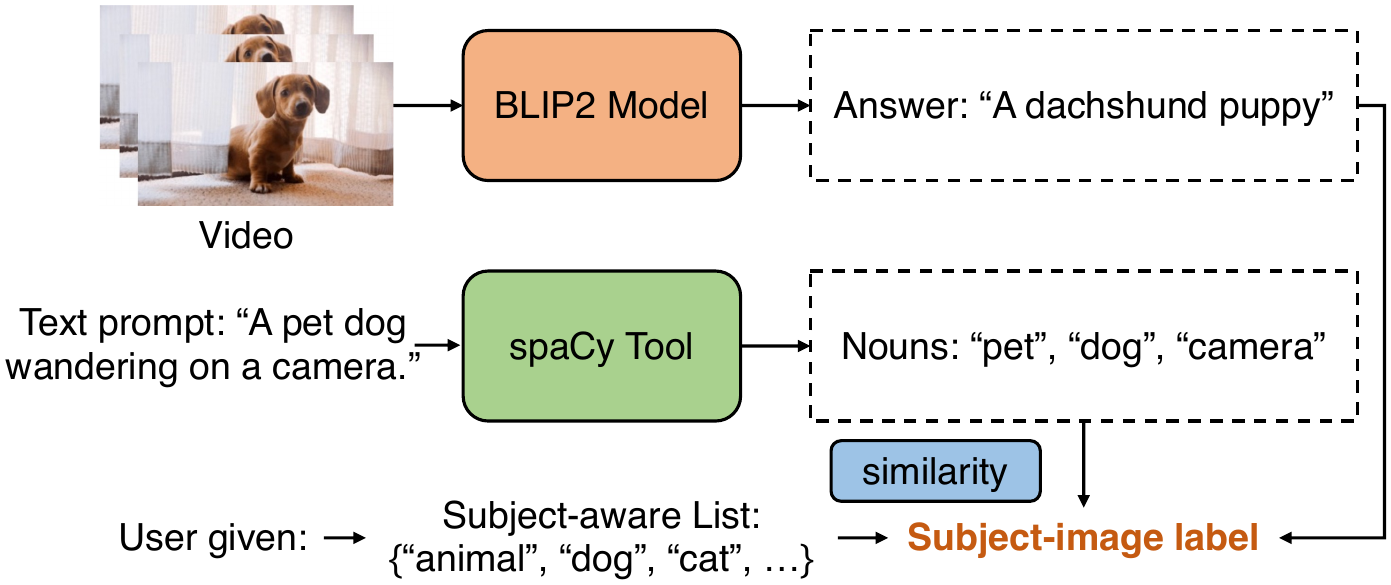}
    \caption{The overview of the method for training data processing.}
    \label{fig: data processing}
\end{figure}

Our proposed MagDiff is a tuning-free method that needs training on the video dataset, which encompasses paired text-video data along with their corresponding segmentation masks.
Due to the lack of existing suitable training data, we propose an automated data processing methodology that segments the subject within a video to construct the requisite training set, shown in Fig.~\ref{fig: data processing}.

We first analyze the existing datasets carefully to get the subject in the image. 
Due to the low quality of the existing dataset, we find that the misalignment between the video data and the caption makes it difficult to get the correct mask area of the subject.
To address this issue, we utilize BLIP-2~\cite{li2023blip2} to generate more precise captions. Specifically, given a video containing $m$ frames $\mathcal{V}=\left\{v_{i} \mid i \in[1, m]\right\}$ with its original caption $\mathcal{P}$, we select the first frame $v_1$ and use BLIP-2 to get question-paired answers. We set the question ``\textit{What is the foreground in the picture?}'' for BLIP-2 and get the answer $\mathcal{A}$. At the same time, the origin prompt $\mathcal{P}$ is handled with spaCy tool to extract nouns $\mathcal{S}=\left\{s_{i} \mid i \in[1, n]\right\}$. To ensure the accuracy of the subject's label, we maintain a user-given subject-aware list $\mathcal{W}$ to prioritize entities for a certain topic, which contains words of some specific domains, such as ``\textit{animal, dog, cat, \dots}''.

We calculate the distance metric to quantify the disparity between the answer $\mathcal{A}$ and each word within the list $\mathcal{W}$. Subsequently, we assess the similarity between the words in $\mathcal{W}$ and each noun in the subject-aware list $\mathcal{S}$, individually. The selection process is defined in \cref{alg:data process.}. Once we get the label of the subject image, we use the SAM-Track~\cite{cheng2023segment} tool to segment the subject mask of each frame. It can ensure the consistency of the segmentations among the frames. Furthermore, we append the final subject-image label to the original video caption, generating an augmented caption. Finally, we merge the video-text pairs with segmentation masks for MagDiff training.

\begin{center}

\begin{minipage}{7cm}

\begin{algorithm}[H]
  \KwIn{$\mathcal{A}$, $\mathcal{W}$, $\mathcal{S}=\left\{s_{i} \mid i \in[1, n]\right\}$.}
  \KwOut{$e$: subject-image label.}
  $scores1$ = List(), $scores2$= List() \\
  \For{$i=0$ \KwTo $length(\mathcal{W})$}{
    scores1.$append$($\mathrm{Similarity(A, \mathcal{W}_i)}$) \\
    \For{$j=0$ \KwTo $length(\mathcal{S})$}{
    scores2.$append$($\mathrm{Similarity(\mathcal{S}_j, \mathcal{W}_i)}$)
    }
  }
  \lIf{$max(scores1)>\theta$}{$e=\mathcal{A}$}
  \lElse{
  $e=max(scores2)$
  }
  \caption{Subject label selection.}
  \label{alg:data process.}
\end{algorithm}

\end{minipage}

\end{center}

\begin{figure}[h]
    \centering
    \captionsetup{type=figure}
    \includegraphics[width=0.6\linewidth]{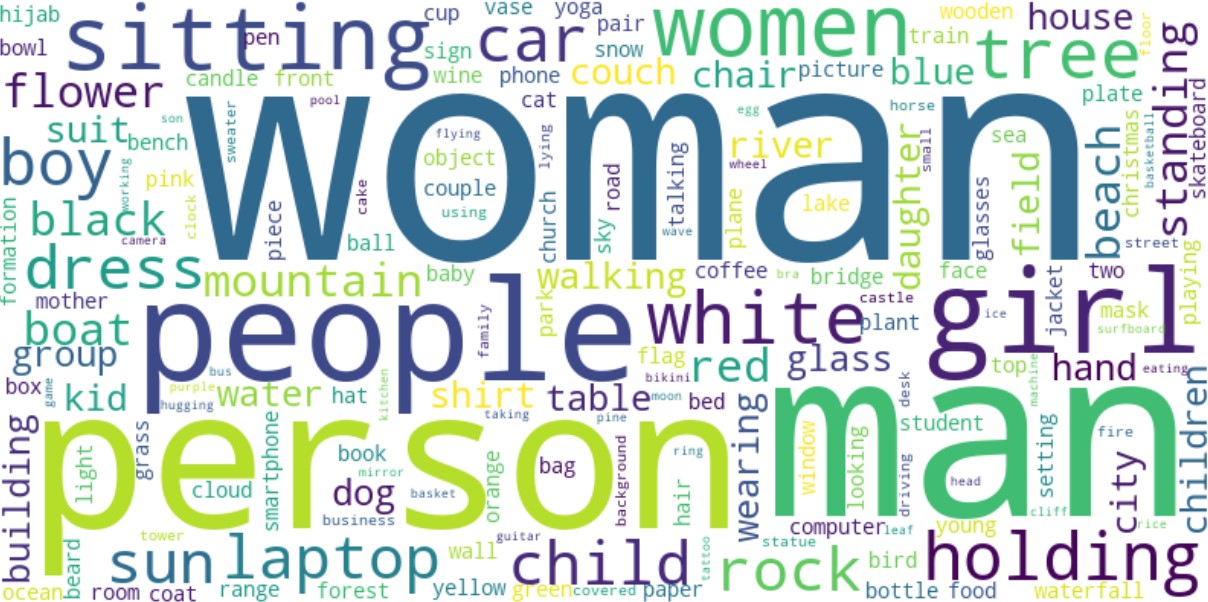}
    \caption{The word cloud of the subject image in training data.}
    \label{fig:entity_word_cloud}
\end{figure}

\begin{figure*}[h]
    \centering
    \captionsetup{type=figure}
    \includegraphics[width=0.95\linewidth]{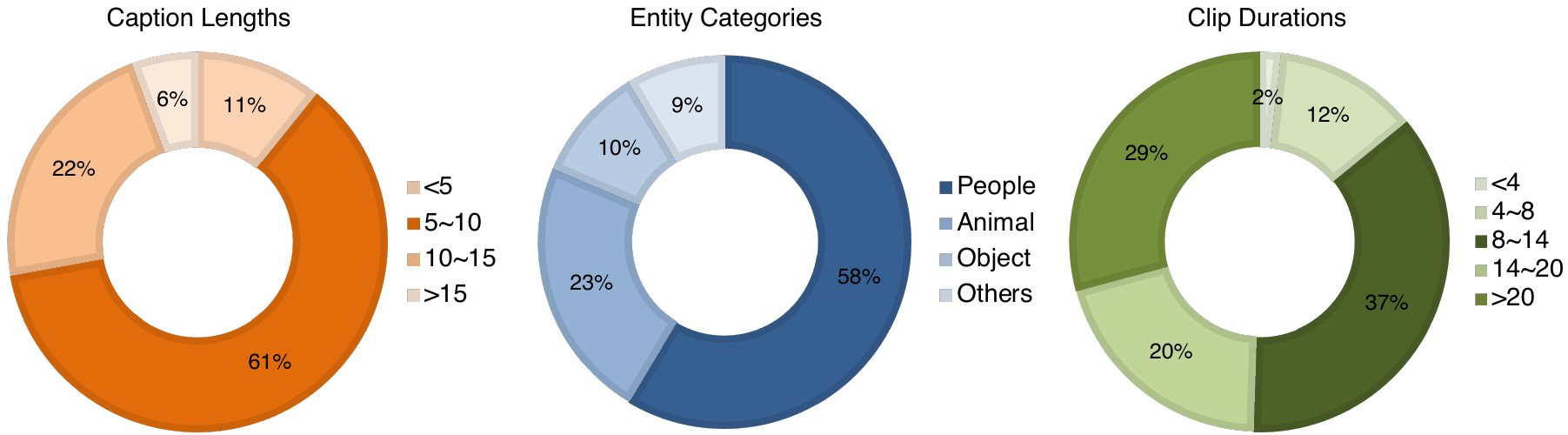}
    \caption{Statistics of caption lengths, entity categories, and clip durations in the training data. Our training data exhibits a diversity of captions and videos with different lengths. The subject regions mainly consist of humans and animals, which offer our training clear boundaries and outstanding dynamic features.}
    \label{fig:data_statistic}
\end{figure*}

Moreover, considering the quality of the existing text-video dataset, we clean up and sample the videos from the Pexel Videos dataset \footnote{https://huggingface.co/datasets/Corran/pexelvideos}. We show the word cloud of subject images in \cref{fig:entity_word_cloud}. In addition, to ensure that the videos are suitable for training, we set three filtering rules to clean up the selected videos:
\textbf{(1)} We filtered out videos with a short side resolution of less than 512. \textbf{(2)} We excluded examples with an entity area ratio of less than 5\% or more than 60\%. \textbf{(3)} We filtered out subject-image labels with no clear meaning.
After filtering, we get 76K videos for training. We simply divide the label classifications into four kinds: person, animal, objects, and others. We show the statistics of caption length, entity categories, and clip durations in \cref{fig:data_statistic}. The dataset contains abundant kinds of subjects, such as \textit{woman, man, person, tree, dog}, and so on. 
Besides, we provide some examples of the training data in \cref{fig:training_data_examples}

\section{Model and Experiments Details}
\label{sec: model and experiments details}
In this section, we introduce the additional model details and experimental setting details. Note that figure and table references with numerical indices pertain to those within the regular \textit{conf.} paper, while those with alphabetical indices refer to the supplementary materials.

\paragraph{Model details.} Our MagDiff is built based on the VidRD~\cite{gu2023reuse} model, which is a text-to-video generation framework trained on 5.3M video-text data. Following VidRD model, we use a regularized autoencoder to compress the original pixels into latent space to save computation and memory. The autoencoder contains an encoder $\mathcal{E}$ for encoding pixel features $\mathbf{x}$ into latent features $\mathbf{z}$ and a decoder $\mathcal{D}$ for decoding $\mathbf{z}$ back to $\mathbf{x}$. We employ the autoencoder which is pre-trained by reconstruction loss. In the training stage, the parameters of the autoencoder are frozen. Besides, in the High-Fidelity Alignment module (HFA), each frame is cropped to the size $384 \times 384$, $320 \times 320$, and $256 \times 256$ (width $\times$ height) for training.
In the Adaptive Prompts Alignment (APA) module, we merge three kinds of resolutions to $256 \times 256$.

\begin{figure*}[h]
    \centering
    \captionsetup{type=figure}
    \includegraphics[width=0.6\linewidth]{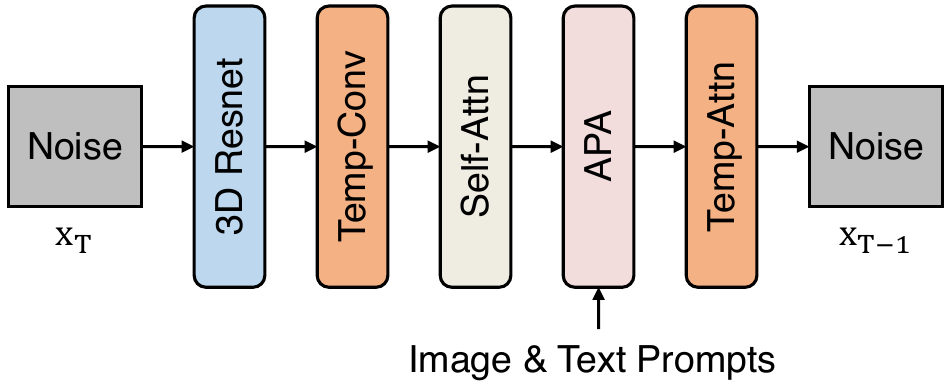}
    \caption{The structure of the U-Net. The noise $\rm{x_T}$ in $\rm{T}$ timestep is denoised into $\rm{x_{T-1}}$.} 
    \label{fig: unet-structure}
\end{figure*}

We provide the structure of the U-Net in Fig.~\ref{fig: unet-structure}. In each denoising step, the noise is processed by 3D Resnet, temporal layers, self-attention layer, and our proposed APA module. The temporal layers include two types of structure: the \textit{Temp-Conv} and the \textit{Temp-Attn}. The \textit{Temp-Conv} represents 3D convolution layers and \textit{Temp-Attn} denotes the temporal attention layers. These temporal layers are injected into the image U-Net structure to learn the action motion and temporal features from video data. 
We load the pre-trained parameters of \textit{Temp-Conv} and \textit{Temp-Attn} from VidRD model. The newly added key and value in the APA module and projects in the HFA module are randomly initialized.
For efficient training, only part of the U-Net network layers are trainable. In Fig.~\ref{fig: unet-structure}, we train the two temporal layers, the self-attention layer, and the APA module, since the Resnet block is frozen. 

Additionally, although our MagDiff can achieve both video generation and editing tasks, we only train the MagDiff once. During training, we only use the first image with its mask as the input for the HFA module and as the subject-image prompt for the APA module. It is not necessary to train the model individually for the video editing task. During video editing inference, we can directly replace the first image with the mask to the whole video with masks for each frame. This effect comes from our designed SDA module, which allows the MagDiff to pay attention to each frame of the input during the denoising process. 
In the generation task, the network generates the frames that the corresponding reference images are masked, thereby achieving content generation by text prompt.
In editing tasks, the network focuses on the areas that need to be edited while preserving the subjects of each reference image, thereby achieving content editing. 
The once-trained and tuning-free inference for two tasks makes our proposed MagDiff more practical.

\paragraph{Experimental setting details.} Here, we introduce the details of our experiments, including detailed settings and further analyses. For comparison of FVD and IS metrics on MSR-VTT and UCF-101, we get the first frame of the video in the two datasets and segment the subject's mask as the input of our model. 
We evaluate all the methods in a tuning-free manner for fair comparison.
It is important to note that, due to the misalignment between the image content and the subject's label, about 0.4\% of the videos cannot be segmented into subjects based on the label in the two datasets. For these examples, we treat the entire video frame as the subject, with the mask set to null. To maintain the fairness of the compared ``text+image-to-video'' methods~\cite{guo2023animatediff, zhang2023i2vgen-xl}, we input the whole image as the condition into these methods for evaluation.
When evaluating the MagDiff on DreamBooth~\cite{ruiz2023dreambooth} dataset, we directly use the images to segment the subjects.

For comparison of the video editing task, we sample eight frames for one video as the input of HFA and APA modules. For each frame in one video, we segment the subject with the same label using the SAM-Track tool. We compare two kinds of editing methods, including fine-tuning methods~\cite{wu2023tune_a_video, qi2023fatezero} and tuning-free method~\cite{brooks2023instructpix2pix}. In a tuning-free manner, our MagDiff has a significant improvement.

For comparison of human evaluations, we carefully select 15 different images and write corresponding text prompts to generate videos, covering diverse scenes, styles, and objects. When doing the user study, 34 volunteers are asked to rank the video quality, text-prompt, and image-prompt alignment from one to five.

In ablation studies, we analyze three alignments in Table 5. We attempt to decouple and individually validate HFA, APA, and SDA modules within MagDiff. ``MagDiff w/o SDA'' represents we do not use the mask of the subject image (indicated by $\rm{\times}$ in the table). Because MagDiff needs the reference image as the input in the HFA module and APA module to introduce the image information, we cannot remove the HFA and the APA at the same time. 
In Fig.5, we test our MagDiff with the image prompt and the text prompt. We employ the image prompts with their corresponding masks (not shown in the figure), \textit{i.e., the Doraemon, the volcano, the panda, and the dog}.
In Fig.6, ``MagDiff w/o SDA'' denotes that the mask is not used and the whole reference image is put into the model. ``MagDiff w/o APA'' and ``MagDiff w/o HFA'' are both tested under the condition of having the SDA module (with mask in the image), representing the scenarios where only the HFA module or the APA module exists, respectively.

\section{More Visualizations}
\label{sec: supple_more_visualizations}
To demonstrate the superiority of our proposed MagDiff, we provide more visualizations of video generation (in Fig.~\ref{fig: case_video_generation}) and video editing (in Fig.~\ref{fig: case_video_editing}).

\section{Limitations}
\label{sec: supple_limitation}
Our proposed MagDiff exhibits remarkable capabilities in preserving the fidelity of the subject image and enough alignment between the image prompt and the text prompt. However, because our model employs the diffusion model as the backbone, it is computationally intensive and time-consuming, especially when dealing with large images. It may also not be suitable for all types of images, such as low-contrast or noisy images. These challenges indicate potential directions for future research, such as efficient inference and model robustness.

\begin{figure*}[t]
    \centering
    \captionsetup{type=figure}
    \includegraphics[width=1.0\linewidth]{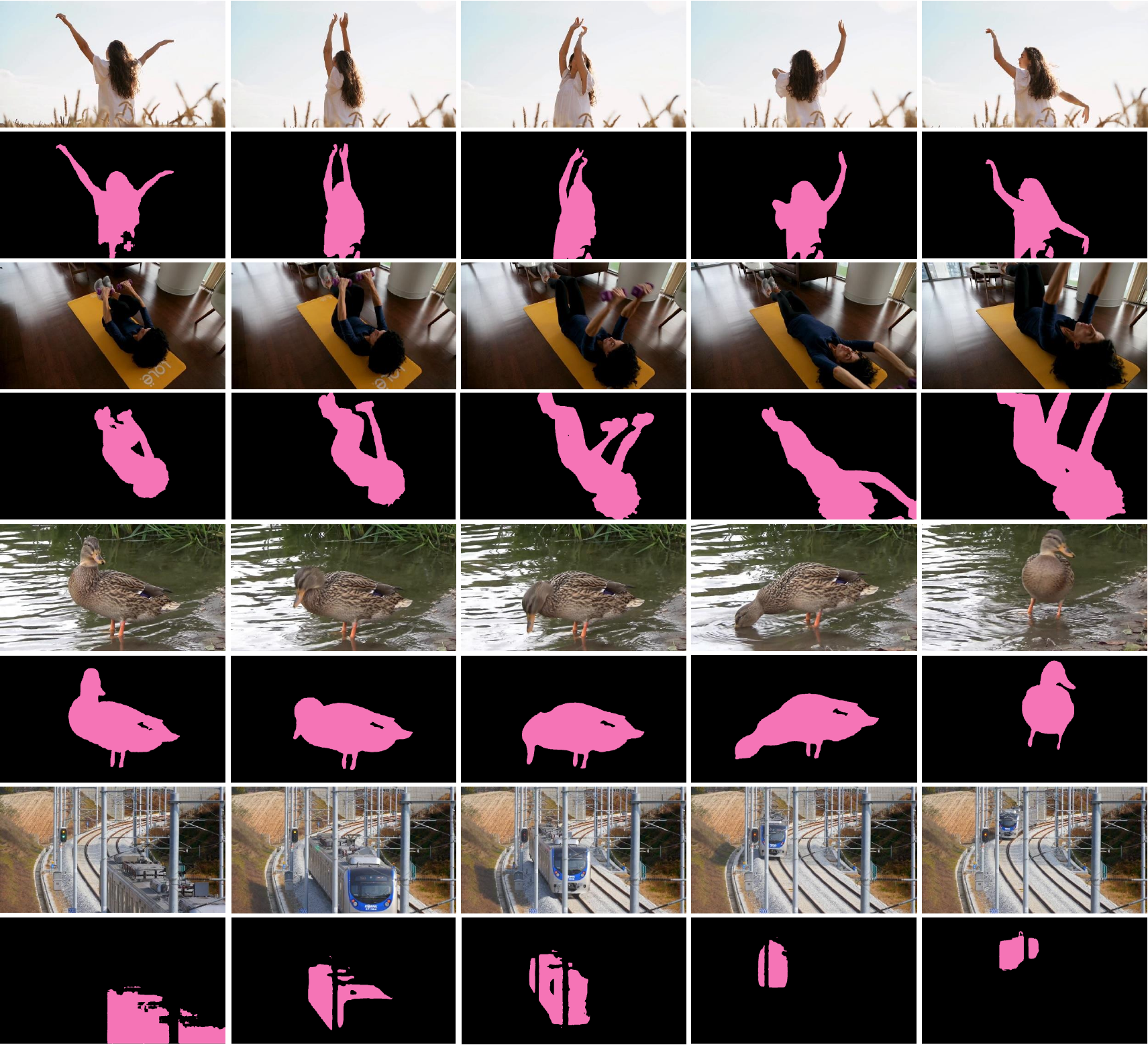}
    \caption{Examples of training data. In each case, the first row is raw video frames and the second row is the subject's mask.}
    \label{fig:training_data_examples}
\end{figure*}

\begin{figure*}[t]
    \centering
    \captionsetup{type=figure}
    \includegraphics[width=1.0\linewidth]{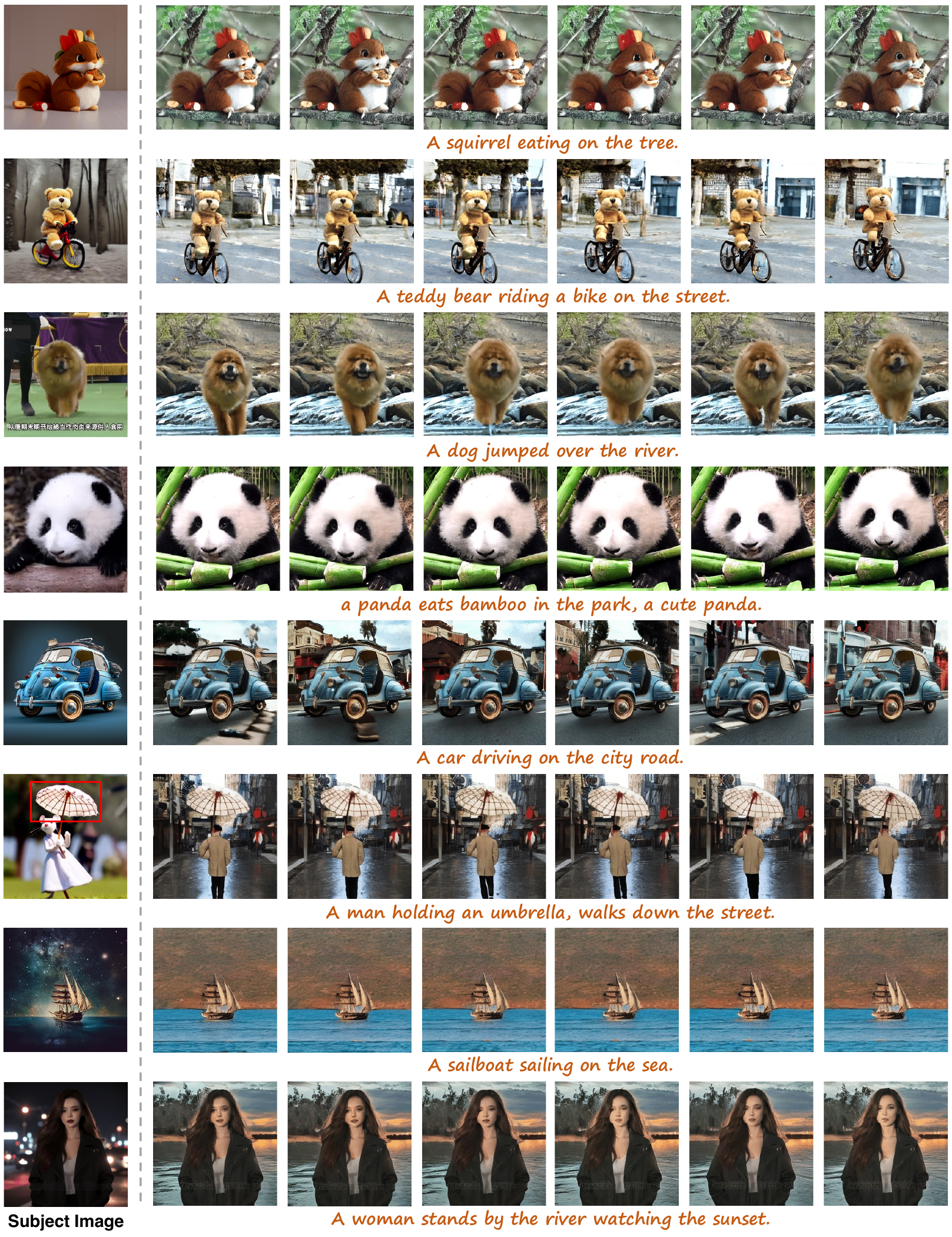}
    \caption{More qualitative generated examples of video generation that conditioned on subject-image prompts. The first column shows the reference subject images.}
    \label{fig: case_video_generation}
\end{figure*}

\begin{figure*}[t]
    \centering
    \captionsetup{type=figure}
    \includegraphics[width=1.0\linewidth]{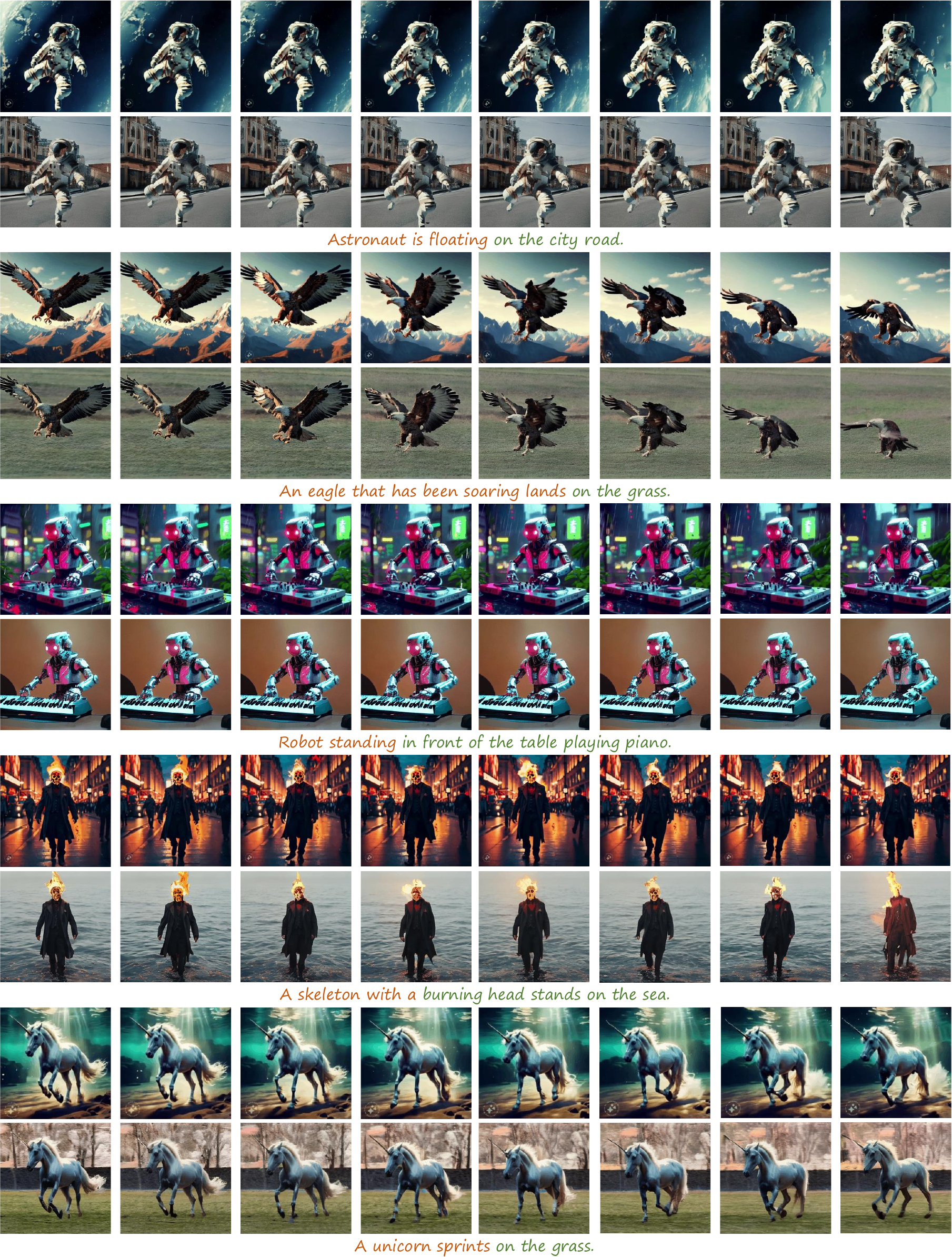}
    \caption{Qualitative results of video editing. In each case, the first row shows the original video and the second row displays the editing result of MagDiff according to the text prompts. We mainly concentrate on editing the green words specified in the prompt.}
    \label{fig: case_video_editing}
\end{figure*}

\clearpage

\end{document}